\documentclass[letterpaper, 10 pt, conference]{ieeeconf}

\IEEEoverridecommandlockouts
\usepackage[caption=false, font=footnotesize]{subfig}
\usepackage[acronym]{glossaries}
\usepackage{booktabs}
\usepackage{textcomp}

\usepackage{graphicx}
\usepackage{subcaption}   % better than subfigure
\usepackage{xcolor}
\usepackage{epsfig} % for postscript graphics files
\usepackage{graphics} % for pdf, bitmapped graphics files
\usepackage{times} % assumes a new font selection scheme installed
\usepackage{amsmath} % assumes amsmath package installed
\usepackage{booktabs, multirow}
\usepackage{soul}% for underlines
\usepackage{changepage,threeparttable} % for wide tables
\usepackage{amssymb}  % assumes amsmath package installed
\usepackage{graphicx}
\usepackage{lipsum}  
\usepackage{amssymb}
\usepackage{array}
\usepackage[nosort,noadjust]{cite}

\usepackage{float}
\usepackage{microtype}
\usepackage{xargs}
\usepackage{booktabs}
\usepackage{bm}
\usepackage{dsfont}

\usepackage[hidelinks]{hyperref}
\usepackage[capitalise]{cleveref}
\usepackage{float}

\definecolor{myred}{HTML}{ff3030}

\title{\LARGE \bf 
Tilt-X: Enabling Compliant Aerial Manipulation through a
Tiltable-Extensible Continuum Manipulator}

\author{Anuraj Uthayasooriyan\textsuperscript{1,2}, Krishna Manaswi Digumarti\textsuperscript{1,2}, Jack Breward\textsuperscript{2}, \\ Fernando Vanegas\textsuperscript{1,2}, Julian Galvez--Serna\textsuperscript{3}, Felipe Gonzalez\textsuperscript{1,2} % <-this % stops a space
\thanks{The authors wish to acknowledge the continued support from the Queensland University of Technology (QUT) through the Centre for Robotics, the support of the Research Engineering Facility (REF) team  and the senior technicians  Steven Bulmer, Benjamin Brownlee, and Amir Moghaddam at QUT for the provision of expertise and research infrastructure, and logistics help from Pirunthan Keertinathan in enablement of this project. Google Gemini was used to edit image backgrounds.}% <-this % stops a space
\thanks{\textsuperscript{1}QUT Centre for Robotics, Queensland University of Technology, Brisbane, QLD Australia 4000, \textsuperscript{2}School of Electrical Engineering and Robotics, Queensland University of Technology, Brisbane, QLD Australia 4000, \textsuperscript{3}REF,Queensland University of Technology, Brisbane, QLD Australia 4000
{\tt\small uanuraj@hotmail.com} , {\tt\small uthayaso@qut.edu.au}}%
}

\usepackage{amsmath}
\usepackage{amssymb}
\usepackage{accents}
\usepackage{bm}
\usepackage{xifthen}
\usepackage{color}
\usepackage{fancyvrb}
\usepackage{graphicx,scalerel}

\newcommand{\ba}{\begin{eqnarray}}
\newcommand{\ea}{\end{eqnarray}}

\newcommand{\presup}[1]{\,{}^{\scriptscriptstyle #1}\!}

\newcommand{\pose}[1][ZZZZ]{\ifthenelse{\equal{#1}{ZZZZ}}{}{\presup{#1}}{\mathbf{\xi}}}
\newcommand{\estpose}[1][ZZZZ]{\ifthenelse{\equal{#1}{ZZZZ}}{}{\presup{#1}}{\mathbf{\hat{\xi}}}}
\newcommand{\hpose}[1][ZZZZ]{\ifthenelse{\equal{#1}{ZZZZ}}{}{\presup{#1}}{\hat{\mathbf{\xi}}}}
\newcommand{\posedot}[1][ZZZZ]{\ifthenelse{\equal{#1}{ZZZZ}}{}{\presup{#1}}{\mathbf{\nu}}}

\newcommand{\q}[1][ZZZZ]{\ifthenelse{\equal{#1}{ZZZZ}}{}{\presup{#1}}{\mathring{q}}}

\DeclareMathAlphabet{\mathitbf}{OML}{cmm}{b}{it}
\newcommand{\twist}[2][ZZZZ]{\ifthenelse{\equal{#1}{ZZZZ}}{}{\presup{#1}}{\mathcal{S}}}
\renewcommand{\vec}[2][ZZZZ]{\ifthenelse{\equal{#1}{ZZZZ}}{}{\presup{#1}}{\mathitbf{#2}}}

\newcommand{\hvec}[2][ZZZZ]{\ifthenelse{\equal{#1}{ZZZZ}}{}{\presup{#1}}{\tilde{\vec{#2}}}}
\newcommand{\obvec}[2][ZZZZ]{\ifthenelse{\equal{#1}{ZZZZ}}{}{\presup{#1}}\rlap{${\overbridge{\phantom{$\vec{#2}$}}}$}\vec{#2}}
\newcommand{\evec}[2][ZZZZ]{\ifthenelse{\equal{#1}{ZZZZ}}{}{\presup{#1}}{\hat{\vec{#2}}}}
\newcommand{\bvec}[2][ZZZZ]{\ifthenelse{\equal{#1}{ZZZZ}}{}{\presup{#1}}{\bar{\vec{#2}}}}

\newcommand{\dvec}[2][ZZZZ]{\ifthenelse{\equal{#1}{ZZZZ}}{}{\presup{#1}}{\dot{\vec{#2}}}}
\newcommand{\ddvec}[2][ZZZZ]{\ifthenelse{\equal{#1}{ZZZZ}}{}{\presup{#1}}{\ddot{\vec{#2}}}}

\newcommand{\mat}[2][ZZZZ]{\ifthenelse{\equal{#1}{ZZZZ}}{}{\presup{#1}\,}{{\boldsymbol #2}}}
\newcommand{\dmat}[2][ZZZZ]{\ifthenelse{\equal{#1}{ZZZZ}}{}{\presup{#1}\,}{{\dot{\boldsymbol #2}}}}
\newcommand{\emat}[2][ZZZZ]{\ifthenelse{\equal{#1}{ZZZZ}}{}{\presup{#1}\,}{\hat{\boldsymbol#2}}}
\newcommand{\matfn}[3][ZZZZ]{\ifthenelse{\equal{#1}{ZZZZ}}{}{\presup{#1}}{{\mat{#2}}\left(#3\right)}}
\newcommand{\Rt}[2][ZZZZ]{\ifthenelse{\equal{#1}{ZZZZ}}{}{\presup{#1}}{{\bf R}\left(#2\right)}}

\newcommand{\point}[2][ZZZZ]{\ifthenelse{\equal{#1}{ZZZZ}}{}{\presup{#1}}{\mathbf{\mathrm{#2}}}}

\newfont{\School}{pncr}
\newfont{\eightTR}{pncr at 8pt}
\usepackage{color}
\usepackage{fancyvrb}
\fvset{formatcom=\color{blue},fontseries=c,fontfamily=courier,xleftmargin=4mm,commentchar=!}

\DefineVerbatimEnvironment{Code}{Verbatim}{formatcom=\color{blue},fontseries=c,fontfamily=courier,fontsize=\footnotesize,xleftmargin=4mm,commentchar=!}

\DefineVerbatimEnvironment{CodeSmall}{Verbatim}{formatcom=\color{blue},fontseries=c,fontfamily=courier,fontsize=\scriptsize,xleftmargin=1mm,commentchar=!}

\DefineVerbatimEnvironment{CodeNum}{Verbatim}{numbers=left,numbersep=4pt,formatcom=\color{blue},fontseries=c,fontfamily=courier,fontsize=\footnotesize,xleftmargin=4mm}

\newcommand{\model}[1]{\index{code}{#1@\textit{#1}}\ifthenelse{\boolean{draft}}{{\color{green}\Verb+#1+}}{\Verb+#1+}}
\newcommand{\block}[1]{\ifthenelse{\boolean{draft}}{{\color{green}\Verb+#1+}}{\textsf{#1}}}

\newcommand{\func}[2][ZZZZ]{\ifthenelse{\equal{#1}{ZZZZ}}{\index{code}{#2}}{\index{code}{#1}}\ifthenelse{\boolean{draft}}{{\color{green}\Verb+#2+}}{\Verb+#2+}}

\newcommand{\methodb}[2]{\index{code}{#1@\textbf{#1}!.#2}\ifthenelse{\boolean{draft}}{{\color{magenta}\Verb+#1.#2+}}{\Verb+#1.#2+}}
\newcommand{\method}[2]{\index{code}{#1@\textbf{#1}!.#2}\ifthenelse{\boolean{draft}}{{\color{magenta}\Verb+#2+}}{\Verb+#2+}}
\newcommand{\class}[1]{\index{code}{#1@\textbf{#1}}\ifthenelse{\boolean{draft}}{{\color{cyan}\Verb+#1+}}{\Verb+#1+}}
\newcommand{\property}[1]{\index{property}{#1}\ifthenelse{\boolean{draft}}{{\color{cyan}\Verb+#1+}}{\Verb+#1+}}

\begin{document}

\makeatletter
\let\@oldmaketitle\@maketitle
\renewcommand{\@maketitle}{\@oldmaketitle
\centering
\includegraphics[width=17.5cm]{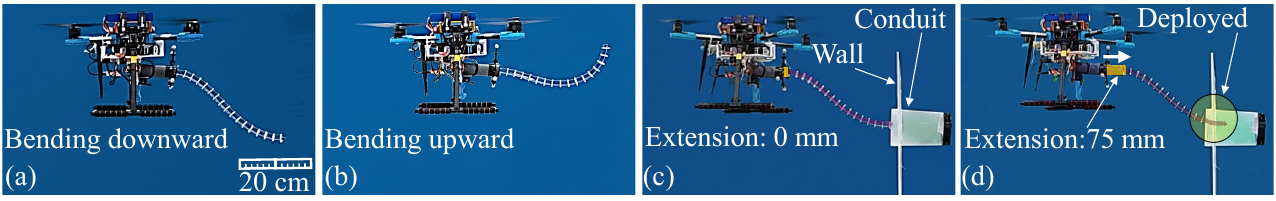}
\vspace{-0.1cm}
\captionof{figure}{A demonstration of the capability of the Tilt-X system that is capable of bending (a,b), tilting and extending (c,d) airborne continuum manipulator. This dexterity makes our system suitable for inspection and deployment tasks.}
\label{fig:main_figure}
}
\makeatother

\maketitle
\setcounter{figure}{1}
\renewcommand{\thefigure}{\arabic{figure}}

\begin{abstract}

Aerial manipulators extend the reach and manipulation capabilities of uncrewed multirotor aerial vehicles for inspection, agriculture, sampling, and delivery. Continuum arm aerial manipulation systems offer lightweight, dexterous, and compliant interaction opportunities. Existing designs allow manipulation only below the UAV which restricts their deployability in multiple directions and through clutter. They are also sensitive to propeller downwash. Addressing these limitations, we present Tilt-X, a continuum arm aerial manipulator that integrates a tilting mechanism, a telescopic stage, and a cable-driven continuum section. We present its design and kinematic model and validate it through flight demonstrations. Tilt-X enables a volumetric workspace with up to 75 mm extension and planar orientations between 0° to 90°. Experiments comparing end effector pose with and without downwash quantitatively measure its accuracy, providing critical evidence to guide the design and control of reliable aerial manipulators. Results show stabilisation of end effector pose as the manipulator extends out of the propeller influence zone. 
\end{abstract}

% \begin{IEEEkeywords}

% \end{IEEEkeywords}

\section{Introduction}
An aerial manipulator refers to a robotic arm mounted on a multirotor uncrewed aerial vehicle (UAV) that performs manipulation tasks in flight. Applications include industrial and high altitude inspection, precision agriculture, environmental sampling, and payload delivery. They are well suited for hazardous environments and accessing areas beyond direct observation \cite{morton2016development}.  Conventional systems with rigid-link arms offer high load capacity and agility  but lack the deployment flexibility, dexterity, and compliance often required in these scenarios \cite{ollero2021past,jalali2022aerial}. Continuum Arm Aerial Manipulation Systems (CAAMS), which integrate a continuum manipulator (CM) with a UAV, address these limitations and are lightweight, dexterous, compliant, and maneuverable solutions \cite{jalali2022aerial, peng2025dexterous}.

Applications of CAAMS in industrial surface inspection, for example, require deployment of a CM through confined passages or perform planar operations requiring consistent end-effector contact \cite{hamaza2018adaptive}. Conduit inspection, environmental sampling, and fruit picking require high maneuverability \cite{peng2025dexterous}. 

Research on CAAMS remains at an early stage. A bench-based single-section CM was proposed in \cite{zhao2022modular}, with analysis focused on its mechanical properties and load handling capacity. \cite{chien2021kinematic,chien2023design} introduced  two-section constant-length CM driven by a ground-based tethered tendon control system, limiting system capability. A CAAMS with a three-section CM for aerial compliant grasping and industrial inspection, exploring both design-based solutions and control strategies was presented in \cite{peng2023aecom,peng2025dexterous}. Eye-in-hand visual servoing control has been explored by \cite{amiri2025high}. 

Challenges in utilizing CAAMS arise from both the operational context and the manipulator design. While UAVs provide overall translational motion, accurately deploying arms in cluttered environments is difficult and prone to collisions. In addition, the soft structure of CMs increases sensitivity to propeller downwash, where airflow, ground, or wall effects \cite{david2024ground, orjales2021towards} can reduce end-effector accuracy, increase actuator load, and shorten the endurance of CAAMS.

One approach to improve deployment and maneuverability by design is to increase the number of bending sections, but this increases the complexity of modeling and control \cite{peng2025dexterous, peng2023aecom}. A constant-length CM mounted beneath the UAV does not have sufficient workspace flexibility due to hardware constraints, particularly the propeller layout.  Variable-length CMs enhance deployment by extending or retracting along the centerline \cite{mishra2017simba,nguyen2015tendon,wang2021design}. Currently, most non-tethered designs in the literature, are capable of manipulation only below the UAV, with the vehicle hovering above the region of interest. However, in practice, the target may also lie beside or above the UAV. To overcome these limitations, this paper presents a CAAMS with a tiltable and extensible mechanism, that expands the reachable workspace. 

By virtue of being compliant, CAAMS can be susceptible to the downwash from propellers and wind effects in the vicinity of obstacles or platforms. Previous research on modeling and control of CAAMS dynamics has primarily relied on simulations, often assuming values for propeller downwash disturbances \cite{hashemi2023robust, ghorbani2023dual, samadikhoshkho2020modeling}. However, experimental characterization of its impact on end-effector positioning accuracy is lacking. Addressing this gap through empirical analysis is essential for developing reliable CAAMS control for precise manipulation in real-world scenarios.

We present Tilt-X (\cref{fig:main_figure}), a novel design that integrates a tilting mechanism, telescopic stage, and cable-driven continuum section. The design enables CM orientation from 0\textdegree (vertically down) to 90\textdegree (horizontally forward) with extensibility. Our design can easily be adopted for multi-section CM designs. We first compare the workspace predicted by the kinematic model with experimental results. We then evaluate the influence of propeller downwash on end-effector pose across multiple tilt angles, extensions, orientations, and bending configurations, including tests near ground and wall surfaces. Results show that Tilt-X mitigates oscillation effects induced by downwash during sideways manipulation.

The contributions of this work are as follows.

\begin{enumerate}
        \item  A novel design and prototype for CAAMS along with a real-flight demonstration of Tilt-X.
        \item  A presentation of kinematics, simulated workspace, and experimental comparison of the Tilt-X with its model. 
        \item An analysis of downwash effects on end-effector pose under free-flight, ground, and wall conditions.
\end{enumerate}

\section{Design}
\label{sec:Design}

\begin{figure*}[!h]
    \centering
    \includegraphics[width=\linewidth]{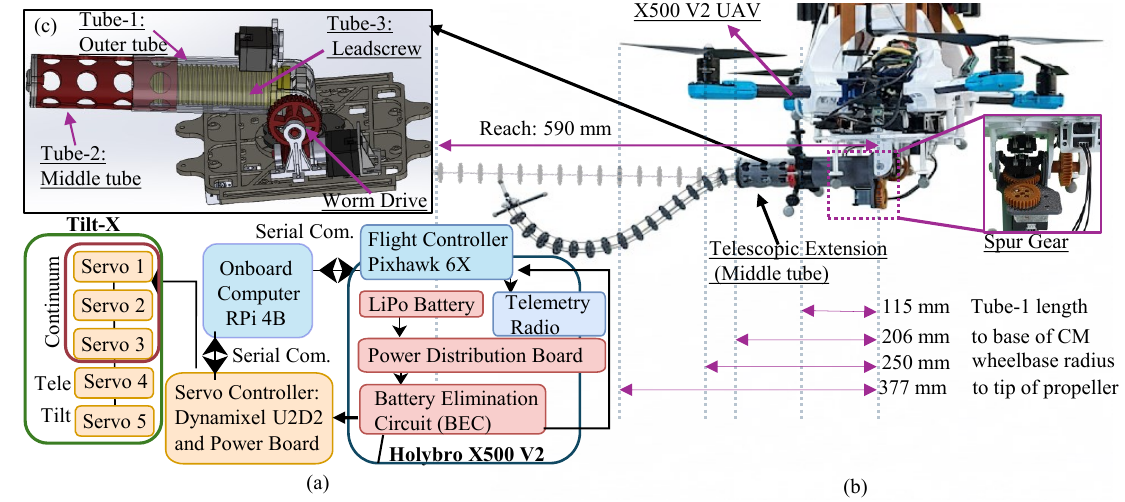}
    \caption{The electrical (a) and mechanical (b) architecture of the Tilt-X system showing the various components. (c) an illustration of the tilt and telescoping mechanism that forms the basis of our design.}
        % \caption{\demo{4}{}}
        \label{fig:System_Architecture}
        \vspace*{-0.1cm}
\end{figure*}

\Cref{fig:System_Architecture} shows the architecture of our CAAMS, which integrates a Holybro X500 V2 quadrotor and the Tilt-X manipulator. The quadrotor uses four 2216-920KV motors (with BLHeli S ESCs and 1045 propellers) controlled by a Pixhawk 6X autopilot. A Raspberry Pi 4B running ROS Noetic on Ubuntu 20.04 serves as the companion computer, communicating with the autopilot via MAVROS/MAVLink and directly controlling Tilt-X. Five Dynamixel XL330-M288-T servos actuate Tilt-X: three for tendons, one for the telescopic stage, and one for tilt control. A Dynamixel U2D2 and a 6V power supply hub enable daisy-chained USB communication between the servos and the companion computer. A ground control station monitors the system and remotely executes precomputed manipulator commands on the onboard computer, sending tendon length variations to position the Tilt-X's end effector.

\subsection{Overall manipulator design}  
We defined the design requirements from applications such as industrial inspection and non-destructive testing, where the manipulator must trace points in space and guide sensors or cameras along complex 3D paths through confined routes \cite{hamaza2018adaptive, hamaza2018towards}. Based on these demands, the following requirements were established. 

\begin{enumerate}
\item Enable controllable bending, translation, and tilting motions while maintaining the center of gravity shift at a minimum.
\item Allow interaction at any angle from horizontally forward to vertically downward relative to the UAV.
\item Extend beyond the sum of the wheelbase radius and the propeller radius of the UAV.
\end{enumerate}

Before developing Tilt-X, we leveraged the UAV’s ability to rotate freely and assumed the workspace could be oriented about the manipulator’s vertical axis. This assumption allowed us to restrict the tilt range to 0\textdegree to 90\textdegree as shown in \cref{fig:Schematic_of_Extensible_CM}. To ensure safe front-facing operations, the manipulator’s minimum horizontal reach was required to exceed the sum of the length of quadrotor's wheelbase radius of 250 mm and the propeller(1045 SF) radius of 127 mm which is 377 mm.  In addition, Tilt-X is designed as a plug-and-play add-on for UAV platforms.

Tilt-X comprises three components: a single-section continuum manipulator, a telescopic stage, and a tilting mechanism. The continuum provides 2-DOF bending in orthogonal planes, while the telescopic and tilting stages add 1-DOF each as axial translation and horizontal rotation. In total, the manipulator offers 4-DOF relative to the UAV hub.
\subsection{Continuum section} 
To demonstrate our proof of concept, we selected a lightweight, single-section cable-driven continuum manipulator design. The CM uses three cables, equally spaced at 120$^\circ$ intervals around the backbone. This design was chosen over a 4-tendon antagonistic one to minimize actuator count and reduce the overall weight \cite{webster2010design, uthayasooriyan2024tendon}. The cross sectional diameter is 20 mm.

We selected segment length and eyelet distance as 0.3, maintaining an eyelet ratio near 0.3 for maximum efficiency, as recommended in \cite{uthayasooriyan2024tendon,li2002design}. The continuum backbone uses superelastic nitinol from NEXMETAL Coorporation. The support disks are 3D-printed in PLA and secured to the backbone with epoxy resin glue from SELLEYS\textsuperscript{\textregistered}. The cables use Power PRO microfilament Spectra fiber braided threads \cite{amanov2021tendon}. Although we demonstrate our concept with a single continuum section, which produces C-shaped bending and 2-DOF, this approach can be extended to multiple sections, providing additional bending degrees of freedom.

\subsection{Telescopic design} 

The telescopic mechanism addresses two key challenges in CAAMS: providing translational motion across orientations from horizontal to vertical, and minimizing center-of-gravity (CoG) shifts during extension. Maintaining CoG stability reduces destabilizing propeller moments \cite{gray2023design}. Previous tendon-driven designs intended for ground operations  translate both actuation and continuum sections on a linear stage \cite{mishra2017simba, wang2021design}. Due to the necesity in aerial manipulator context, Tilt-X fixes the actuation compartment at the UAV hub, limiting CoG shift and enhancing flight stability.

\cref{fig:System_Architecture}(b) shows the telescopic mechanism of three concentric tubes. The innermost hollow leadscrew, driven by a Dynamixel XL330-M288-T servo via a spur gear pair, translates the middle tube carrying the CM. Translation occurs without rotation as grooves on tube 2 engage with rails in tube 1. Tendon length changes from leadscrew translation and tilting are incorporated into the kinematic model in Section~\ref{sec:Model}. Mounting holes at tube 2’s free end allow attachment of cable-driven CMs within specified diameter and provide paths for tendons.

\subsection{Tilting mechanism}The Tilt-X is attached to the UAV via a revolute joint which allows rotation in the xz plane of the CAAMS. The range of this joint is 0\textdegree–90\textdegree (vertically down to horizontally forward) to allow for manipulation in all orientations from below to in front of the UAV. To ensure stability against external moments, the revolute joint is connected to the drive motor via a 3D printed, single start worm gear pair with a gear ratio of 35:1. 

\section{Modeling} 
\label{sec:Model}
We adopt a decoupled kinematic model, treating the CM as a standalone system independent of UAV kinematics \cite{samadikhoshkho2020modeling}. This corresponds to assuming the UAV hovers at a fixed position, with the flight controller compensating for inertia variations, CoG shifts, and inertial forces. Accordingly, Tilt-X forward kinematics are derived as if the CM were attached to a fixed base.

\begin{figure}[ht]
    \centering
    \includegraphics[width=0.35\textwidth]{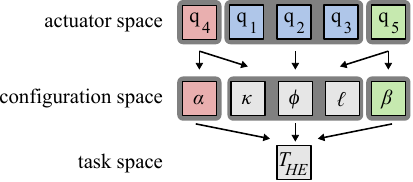}
    \caption{ Forward kinematics map from the actuator-space ($q_1$-$q_5$: motor angles) to configuration-space (curvature $\kappa$, bending plane angle $\phi$, arc length $\ell$, tilt angle $\alpha$, linear extension $\beta$) to task-space pose of the end effector $\mathbf{T}_{HE}$.}
        % \caption{\demo{4}{}}
        \label{fig:Mapping_between_spaces}
        \vspace*{-0.5cm}
\end{figure}

The forward kinematics of the continuum portion of Tilt-X is derived using the constant curvature assumption as in \cite{webster2010design}. We assume that the cables are inextensible and without slack, and ignore moments on the CM. Since our design is unique in that the CM's actuators do not move with the tilting and telescoping portion of the manipulator, the forward kinematics is also unique. \cref{fig:Mapping_between_spaces} shows the relationship between the actuation space, configuration space and the task space (work space). The actuators constantly adjust cable lengths accounting for changes produced by both the tilt and telescopic mechanism. 

\subsection{Mapping configuration space to task space}

\begin{figure}[ht]
    \centering
    \includegraphics[width=\linewidth]{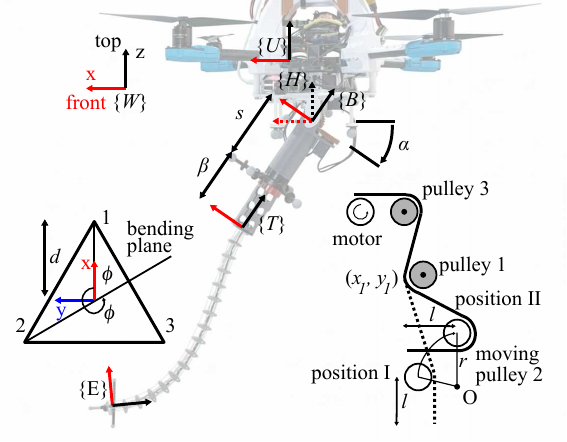}
    \caption{(a) A schematic of the system showing the key coordinate frames - $\{W\}$: world,$\{U\}$: UAV,$\{H\}$: hinge  on the UAV, $\{B\}$: base of the Tilt-X, $\{T\}$: tip of the telescopic mechanism, and $\{E\}$: tip of the continuum section. (b) The position of the cables 1-3 relative to the bending plane on the triangular base. (c) The routing of the cables through a pulley system at the hinge is on the bottom right.}
        % \caption{\demo{4}{}}
    \label{fig:Schematic_of_Extensible_CM}
\end{figure}

In our design, frames are defined as follows. $\{W\}$: world frame, $\{U\}$: UAV frame, $\{H\}$: frame on the UAV where the Tilt-X is attached by a hinge, $\{B\}$: frame at the base of the Tilt-X where it connects to the hinge, $\{T\}$: tip of the telescopic mechanism which serves as the base of the continuum section, and $\{E\}$: frame attached to the tip of the continuum section (\cref{fig:Schematic_of_Extensible_CM}).

The pose of the end effector relative to the world frame can be described through a sequence of homogeneous transformations as

\begin{equation}
\label{eq:WorldToEE}
    \mathbf{T}_{WE} = \mathbf{T}_{WU} \cdot \mathbf{T}_{UH} \cdot \mathbf{T}_{HB} \cdot \mathbf{T}_{BT} \cdot \mathbf{T}_{TE}.
\end{equation}

The position of the UAV in the world frame is assumed to be known. Referring to \cref{fig:Schematic_of_Extensible_CM}, by design, the hinge $\{H\}$ and the UAV's body frame $\{U\}$ are separated only by a translation with no rotation. Therefore, we can write
\begin{equation}
\label{eq:UAVtoHinge}  
\mathbf{T}_{UH} = \mathbf{T}_{t_x}(H_{\text{x}}) \cdot \mathbf{T}_{t_y}(H_{\text{y}}) \cdot \mathbf{T}_{t_z}(-H_{\text{z}}).
% \begin{bmatrix}
% 1 & 0 & 0 & H_{\text{x\_offset}} \\
% 0 & 1 & 0 & H_{\text{y\_offset}} \\
% 0 & 0 & 1 & -H_{\text{z\_offset}} \\
% 0 & 0 & 0 & 1
% \end{bmatrix}
\end{equation}

The origins of frames $\{H\}$ and $\{B\}$ coincide and frame $\{B\}$ rotates about the y-axis of frame $\{H\}$. Thus,
\begin{equation}
\label{eq:HingeToBase}
\mathbf{T}_{HB} = \mathbf{T}_{R_y}(-\alpha).
% \begin{bmatrix}
% \cos\alpha & 0 & \sin\alpha & 0 \\
% 0 & 1 & 0 & 0 \\
% -\sin\alpha & 0 & \cos\alpha & 0 \\
% 0 & 0 & 0 & 1
% \end{bmatrix}
\end{equation}

By design, the tip $\{T\}$ is offset from $\{B\}$ by a fixed distance of $s$=131 mm and a variable distance $\beta \in [0,75]$ mm in the negative $z$ direction. Thus,

\begin{equation}
\label{eq:BaseToTelescope}
\mathbf{T}_{BT} = \mathbf{T}_{t_z}(-(s+\beta)).
% \begin{bmatrix}
% 1 & 0 & 0 & 0 \\
% 0 & 1 & 0 & 0 \\
% 0 & 0 & 1 & -(s+z) \\
% 0 & 0 & 0 & 1
% \end{bmatrix}
\end{equation}

To determine the pose of the end effector $\{E\}$ with respect to the tip $\{T\}$, we follow the process given in \cite{webster2010design}, noting that the orientations of frames in our model are different from theirs. The curve is defined by its curvature $\kappa$, bend plane angle $\phi$ and length $\ell$. Defining $c_\phi = \cos\phi$, $s_\phi = \sin\phi$, $c_{\kappa\ell} = \cos({\kappa\ell})$, and $s_{\kappa\ell} = \sin({\kappa\ell})$, the pose of the end effector in frame $\{T\}$ is

\begin{equation}
\label{eq:TelToEE_homogeneous}
T_{TE} =
\begin{bmatrix}
c_{\kappa\ell} c_\phi & s_\phi & s_{\kappa\ell} c_\phi & \frac{1}{\kappa}(1-c_{\kappa\ell})c_\phi \\
c_{\kappa\ell} s_\phi & -c_\phi & s_{\kappa\ell} s_\phi & \frac{1}{\kappa}(1-c_{\kappa\ell})s_\phi \\
s_{\kappa\ell} & 0 & -c_{\kappa\ell} & -\frac{1}{\kappa}s_{\kappa\ell} \\
0 & 0 & 0 & 1
\end{bmatrix}.
\end{equation}
Combining \cref{eq:HingeToBase} through \cref{eq:TelToEE_homogeneous} gives the relative pose of the end effector with respect to the hinge frame.

%%%%%%%%%%%%%%%%%%%%%%%%%%%%%%%%%%%%%%%%%%%%%%%%%%%%%%%%%%%%%%%%%%%%
%Simulated workspace
%%%%%%%%%%%%%%%%%%%%%%%%%%%%%%%%%%%%%%%%%%%%%%%%%%%%%%%%%%%%%%%%%%%%

\begin{equation}
\label{HingeToEE_final}
\mathbf{T}_{HE} =
\begin{bmatrix}
\mathbf{R}_{HE} & \mathbf{t}_{HE} \\
\mathbf{0}_{1\times 3} & 1
\end{bmatrix}
\end{equation}

where,

\begin{equation}
\label{R_HE}
\mathbf{R}_{HE} =
\begin{bmatrix}
c_\alpha c_{\kappa\ell} c_\phi + s_\alpha s_{\kappa\ell} & c_\alpha s_\phi & c_\alpha s_{\kappa\ell} c_\phi - s_\alpha c_\theta \\
c_{\kappa\ell} s_\phi & -c_\phi & s_{\kappa\ell} s_\phi \\
- s_\alpha c_{\kappa\ell} c_\phi + c_\alpha s_{\kappa\ell} & -s_\alpha s_\phi & -s_\alpha s_{\kappa\ell} c_\phi - c_\alpha c_{\kappa\ell} \\
\end{bmatrix}
\end{equation} 

\begin{equation}
\label{t_HE}
\mathbf{t}_{HE} =
\begin{bmatrix}
\frac{1}{\kappa}  c_\alpha (1-c_{\kappa\ell})c_\phi - \frac{1}{\kappa} s_\alpha s_{\kappa\ell} - s_\alpha (s+\beta)\\
\frac{1}{\kappa} (1-c_{\kappa\ell})s_\phi \\
- \frac{1}{\kappa}  s_\alpha (1-c_\theta)c_\phi - \frac{1}{\kappa}  c_\alpha s_{\kappa\ell} - c_\alpha (s+\beta) \\
\end{bmatrix}
\end{equation}
\cref{HingeToEE_final} maps the configuration space $\left(\kappa, \ell, \phi, \alpha, \beta\right)$ to the task space of the manipulator $\mathbf{T}_{HE}$.

%%%%%%%%%%%%%%%%%%%%%%%%%%%%%%%%%%%%%%%%%%%%%%%%%%%%%%%%%%%%%%%%%%%%%%%%%%%
%Mapping between actuation space and configuration space:
%%%%%%%%%%%%%%%%%%%%%%%%%%%%%%%%%%%%%%%%%%%%%%%%%%%%%%%%%%%%%%%%%%%%%%%%%%%

\subsection{Mapping actuation space to configuration space:}
% \textcolor{red}{use the papers \cite{chien2021kinematic}, \cite{peng2025dexterous} and \cite{mishra2017simba}}\\

The system has five actuators: three to pull cables ($q_1-q_3$), one to rotate the hinge ($q_4$) and one for linear extension ($q_5$). All these actuations cause a change in cable lengths $\Delta l_{i}$ and the relation is described here.

% The corresponding angular positions of the motors are $q_1$, $q_2$ and $q3$.
\cref{fig:Schematic_of_Extensible_CM} depicts the arrangement of cables on a cross section of the manipulator. Cable-1 is located along the X-axis of the continuum section at the base (note the difference from \cite{webster2010design}). Each cable is at a distance $d$ from the centroid. For this specific arrangement, assuming an inextensible backbone with a constant arc length $L$ , we have the following expressions. 
\begin{equation}
    \label{l_q}
\ell(q) = L
\end{equation}

\begin{equation}
\label{BendingPlaneAngle}
\phi(q) = \tan^{-1}\!\left(
\frac{\sqrt{3}\,(l_{2} - l_{3})}{\big(l_{2} + l_{3} - 2l_{1}\big)}
\right)
\end{equation}

\begin{equation}
\label{Kappa}
\kappa(q) = \frac{2}{d\,(l_{1} + l_{2} + l_{3})}
\sqrt{\,l_{1}^{2} + l_{2}^{2} + l_{3}^{2} - l_{1}l_{2} - l_{1}l_{3} - l_{2}l_{3}}
\end{equation}

where $l_{i}$ indicates the length of each cable. The change in length $\Delta l_{CM_i}$ to produce a change in curvature and bend in a desired plane for the continuum portion alone can be described as

\begin{equation}
\label{continuum_tendon_Del_L}
\begin{aligned}
\Delta l_{CM_i} &= -L \kappa d \cos\left( \phi_{i}\right), \\
% \Delta l_{CM_{i=1:3}} &= [\Delta l_{CM_1},\Delta l_{CM_2} ,\Delta l_{CM_3} ]^T
\end{aligned}
\end{equation}

where $\phi_{i}$ is the angular position of the cable with respect to the bending plane (\cref{fig:Schematic_of_Extensible_CM}(b)).
% This means that the ${i}^{th}$  tendon is winding up by $\Delta l_{i}$ to produce a particular curvature ${\kappa}$ at a particular bending plane angle ${\phi}$.

%%%%%%%%%%%%%%%%%%%%%%%%%%%%%%%%%%%%%%%%%%%%%%%%%%%%%%%%%%%%%%%%%%%%
%Telescopic mechanism
%%%%%%%%%%%%%%%%%%%%%%%%%%%%%%%%%%%%%%%%%%%%%%%%%%%%%%%%%%%%%%%%%%%%

The telescopic mechanism provides one linear degree of freedom. Noting $q_5$ as the angle of the motor driving this mechanism, ${P}$ as the pitch of the lead-screw and ${N}_{t}$ as the gear ratio of the spur gear pair which connects the servo motor and the lead-screw, the corresponding changes in length are equal on all three cables and are

\begin{equation}
\label{Delta_L_tele}
\begin{aligned}
\Delta l_{\text{Tele}_{i}} = {\Delta {\beta}} = {P}{N}_{t}{\Delta {q_5}} \\
% \Delta L_{\text{tele}_{i=1:3}} = ({N}_{telescope}({\Delta {\beta}_{mt}}){P})[1,1,1]^T
\end{aligned}
\end{equation}

%%%%%%%%%%%%%%%%%%%%%%%%%%%%%%%%%%%%%%%%%%%%%%%%%%%%%%%%%%%%%%%%%%%%
%Tilting mechanism
%%%%%%%%%%%%%%%%%%%%%%%%%%%%%%%%%%%%%%%%%%%%%%%%%%%%%%%%%%%%%%%%%%%%

The tilt mechanism provides one rotational degree of freedom about the $y$ axis of the hinge at the base of the telescopic mechanism. The tilt ${\alpha}$ is related to the motor angle ${q}_{4}$ through the gear ratio of a one start worm drive ${N}_{w}$.

\begin{equation}
    \label{Tilt_worm}
     {\alpha}=  {{N}_{w}}^{-1}{2\pi q_4}.
\end{equation}

\begin{figure}[h] % use [t] or [h] depending on placement
    \centering
    \includegraphics[width=\linewidth]{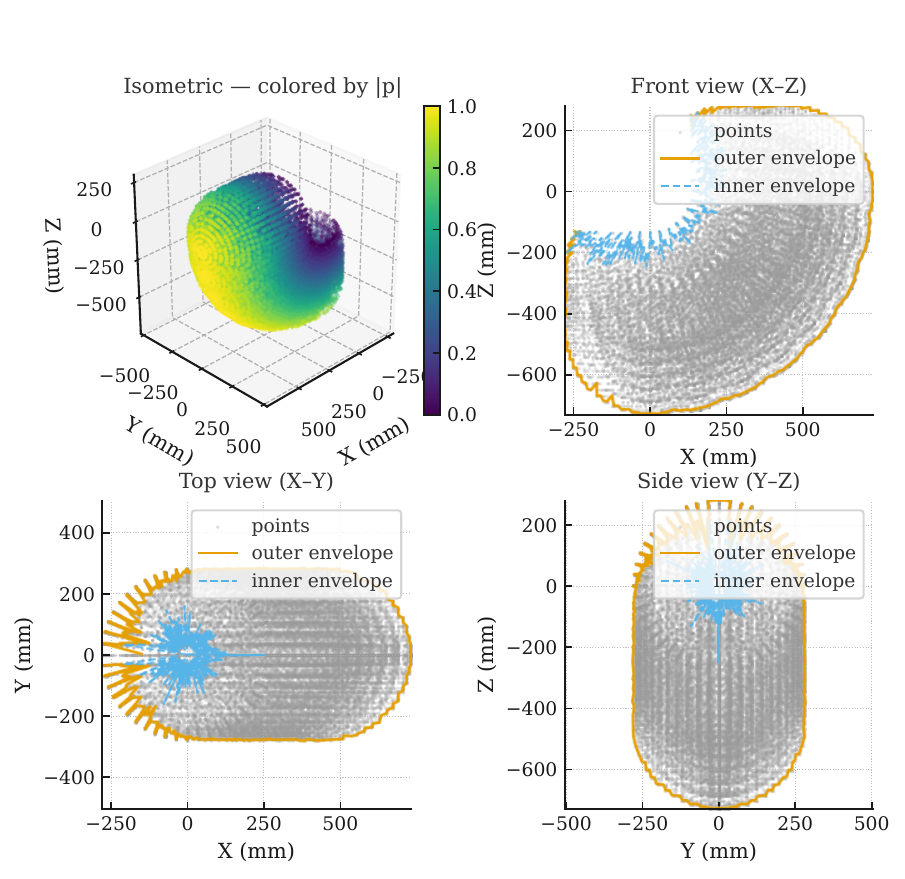}
    \caption{Workspace of the Tilt-X system. 
    (a) Isometric view colored by radial distance. 
    (b) Front view (X–Z). 
    (c) Top view (X–Y). 
    (d) Side view (Y–Z).}
    \label{fig:tdcr_workspace}
\end{figure}

An important consideration is compensating for cable length changes during tilting. To preserve the intended curvature of the continuum section when tilting, the required length change must be taken into account. Cables 2 and 3 are routed through a system of pulleys such that the change in length is negligible given a small pulley radius (2 mm). 

However, cable-1 is routed through an arrangement which induces a change due to the movement of pulley-2 as depicted in \cref{fig:Schematic_of_Extensible_CM}(c). Neglecting the small change in length due to changes in wrapping angle, the change in length due to tilting motion $\Delta{{L}_{tilt}}$ is simply the change in the Eucledean distance between the centers of pulley-1, $c_1=(x_1, y_1)$ and pulley-2, $c_2$=$(r\sin \alpha$, $-r\cos\alpha$) between positions I and II (\cref{fig:Schematic_of_Extensible_CM}c).

\begin{equation}
    \label{TendonRelax}
    {\Delta{{l}_{Tilt} = {C}_{Tilt}(\alpha_{I})- {C}_{Tilt}(\alpha_{II})}}.
\end{equation}
\begin{equation}
    \label{TiltPulley_CableLength}
{C}_{Tilt}(\alpha) = \| c_{2}(\alpha) - c_{1} \| 
% = \sqrt{ \big(x_{1} - r \sin\alpha \big)^{2} + \big(y_{1} + r \cos\alpha \big)^{2} }
\end{equation}

%%%%%%%%%%%%%%%%%%%%%%%%%%%%%%%%%%%%%%%%%%%%%%%%%%%%%%%%%%%%%%%%%%%%
%Actuation of Tendon Motors of Continuum Section
%%%%%%%%%%%%%%%%%%%%%%%%%%%%%%%%%%%%%%%%%%%%%%%%%%%%%%%%%%%%%%%%%%%%

Each motor for the cables is attached to a radius of ${r}_{\text{c-motor}_i}$ and responsible for driving the continuum section. A turn of ${q}_{i}$ translates to a total change in length of ${\Delta l_{i}}$.

\begin{equation}
    \label{Del_L_motor}
    {\Delta l_{i} = {r}_{\text{c-motor}_i}{q}_{i}},
\end{equation}

which accounts for changes in lengths from all sources.

\begin{equation}
\label{Del_L_all}
\Delta l_{i} = \Delta l_{\text{Tilt}_{i}}  + \Delta l_{\text{Tele}_{i}} + \Delta l_{CM_{i}}.
\end{equation}

\begin{figure}[h]
    \centering
    \includegraphics[width=0.45\textwidth]{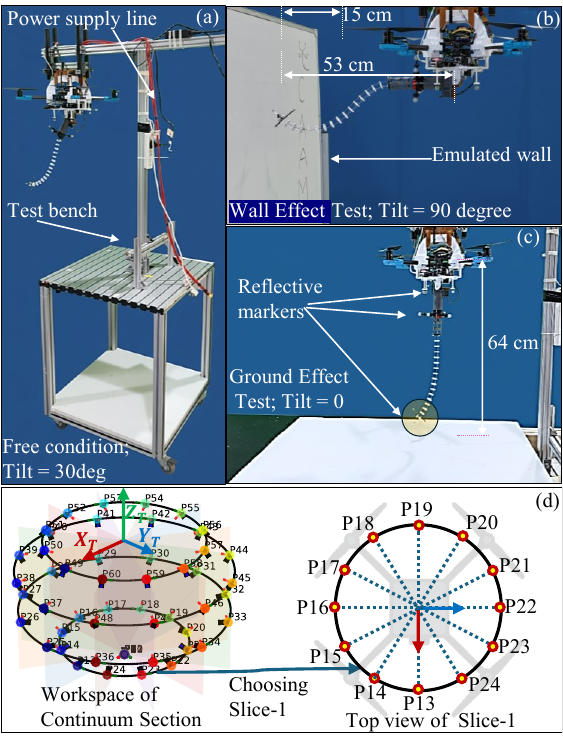}
    \caption{(a) Bench-top experimental setup, (b) setup to evaluate wall effect in which a board is placed next to the UAV, (c) setup to evaluate ground effect in which the UAV approaces close to a board and (d) the location of test points around the UAV used for comparing model with experiments.}
    \label{fig:ExperimentSetup}
\end{figure}

\subsection{Workspace Enhancement}
Using the kinematics model, we analyze the manipulator workspace relative to the UAV. Tilt and telescopic mechanisms address two limitations: (i) lateral tasks require the end effector to exceed the UAV center–propeller span, and (ii) a single-section CM reaches only points on a sphere-like surface centered at its base. Tilt-X instead generates a volumetric workspace, achieved by spanning end-effector poses from non-extended to extended at zero tilt, then sweeping through 0–90°. Unlike backbone extension alone, which enlarges the workspace at the cost of accuracy \cite{grassmann2022fas}, Tilt-X preserves CM integrity. As shown in \cref{fig:tdcr_workspace}, the manipulator has a total reach of 590 mm which exceeds the wheelbase radius and the distance to the tip of the propellers.

\section{Experiments}
\label{sec:Experiment}
\subsection{Experiment Setup}

%%%%%%%%%%%%%%%%%%%%%%%%%%%%%%%%%%%%%%%%%%%%%%%%%%%%%%%%%%%%%%%%
%EXPERIMENT SCOPE
%%%%%%%%%%%%%%%%%%%%%%%%%%%%%%%%%%%%%%%%%%%%%%%%%%%%%%%%%%%%%%%%

We designed two sets of experiments: (i) bench-top workspace evaluation and (ii) flight tests. The bench-top evaluation has two objectives: (a) to compare the manipulator’s workspace with its kinematic model when the propeller is OFF, and (b) to quantify end-effector accuracy by comparing the cases with propellers ON and OFF, including the effects of ground and wall without any influence from the UAV's autopilot. For the propellers-ON case, the rotors were operated at maximum throttle to generate the highest possible downwash assuming that this provides highest possible disturbance in the worst case scenario. \cref{tab:experimentDesign}, summarizes the cases evaluated.

%%%%%%%%%%%%%%%%%%%%%%%%%%%%%%%%%%%%%%%%%%%%%%%%%%%%%%%%%%%%%%%%
%EXPERIMENT SETUP
%%%%%%%%%%%%%%%%%%%%%%%%%%%%%%%%%%%%%%%%%%%%%%%%%%%%%%%%%%%%%%%%

\begin{table}[h]
    \centering
    \caption{Conditions in the Bench-top Experiments}
    \label{tab:experimentDesign}
    \begin{tabular}{l l l l l}
        \toprule
        Exp. & Propeller & Influence & Tilt $\alpha$ & Tilt $\alpha$ \\ 
        & ON/OFF & (wall/ ground)& $\beta$ = 0 mm & $\beta$ = 75 mm\\
        \midrule
        1 & OFF & N/A & $0^{\circ}$, $30^{\circ}$, $90^{\circ}$ & $0^{\circ}$, $30^{\circ}$, $90^{\circ}$ \\
        2 & ON  & Free  & $0^{\circ}$, $30^{\circ}$, $90^{\circ}$ & $0^{\circ}$, $30^{\circ}$, $90^{\circ}$ \\
        3 & ON  & Wall & - & $90^{\circ}$ \\
        4 & ON  & Ground & - & $0^{\circ}$, $30^{\circ}$ \\
        \bottomrule
    \end{tabular}
\end{table}

%%%%%%%%%%%%%%%%%%%%%%%%%%%%%%%%%%%%%%%%%%%%%%%%%%%%%%%%%%%%%%%%
%EXPERIMENT STEPS AND DESIGN
%%%%%%%%%%%%%%%%%%%%%%%%%%%%%%%%%%%%%%%%%%%%%%%%%%%%%%%%%%%%%%%%

We mounted the system with Tilt-X and UAV onto a test bench (see \cref{fig:ExperimentSetup}a), with all six UAV degrees of freedom constrained by a locking mechanism. The system was powered using a 12 V DC supply rated at 20 A, exceeding the current requirement of the Holybro 2216 KV920 brushless motors at 12 V. As described in the system architecture, battery elimination circuits (BECs) provided a regulated 6 V supply for the manipulator motors. The test bench was placed in a tracked arena with OptiTrack® motion capture system running Motive software. Reflective markers were attached to the frames $\{U\}$,$\{H\}$,$\{B\}$, $\{T\}$ and $\{E\}$.

The UAV’s  flight controller was remotely connected to QGroundControl flight control software. In addition, a remote controller (RC) link was established through an onboard RC receiver-transmitter. A separate computer maintained a remote connection with the onboard Raspberry Pi, which handled Tilt-X actuation, monitoring, and data logging through ROS.

To evaluate the influence of proximity to a wall, a whiteboard was placed parallel to the propellers at the front, nearly touching the tip of Tilt-X, as shown in \cref{fig:ExperimentSetup}(b). The distance between the propeller tips and the board was around 16 cm. For the ground influence test, a flat board was positioned directly beneath Tilt-X, also near its tip, with a UAV base-to-board distance of 65 cm as shown in \cref{fig:ExperimentSetup}(c).
%%%%%%%%%%%%%%%%%%%%%%%%%%%%%%%%%%%%%%%%%%%%%%%%%%%%%%%%%%%%%%%%
%Figure caption
%%%%%%%%%%%%%%%%%%%%%%%%%%%%%%%%%%%%%%%%%%%%%%%%%%%%%%%%%%%%%%%%

%%%%%%%%%%%%%%%%%%%%%%%%%%%%%%%%%%%%%%%%%%%%%%%%%%%%%%%%%%%%%%%%
%DATA COLLECTION
%%%%%%%%%%%%%%%%%%%%%%%%%%%%%%%%%%%%%%%%%%%%%%%%%%%%%%%%%%%%%%%%

We first sampled the theoretical workspace of the continuum section derived in Section\cref{sec:Model} at four offsets (hereafter referred to as slices) along the continuum section’s -Z axis from its base. For each slice, with bending plane angles ${\phi}$ spaced at 30\textdegree, we precomputed the required tendon lengths using the continuum model. This produced a dataset of end-effector positions, orientations, and tendon lengths for forward-kinematics-based actuation. The sampled points and corresponding bending planes are illustrated in \cref{fig:ExperimentSetup}(d). 

\begin{figure}
    \centering
    \includegraphics[width=\linewidth]{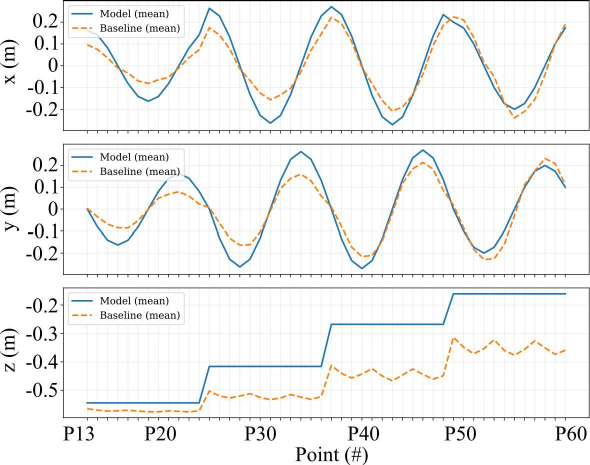}
    \caption{The x,y,z positions of the end effector when only the continuum section is actuated with no tilt present, at an extension of 75 mm.}
    \label{fig:continuumXYZ_positions}
\end{figure}

\begin{figure}[!ht]
    \centering
    \includegraphics[width=\linewidth]{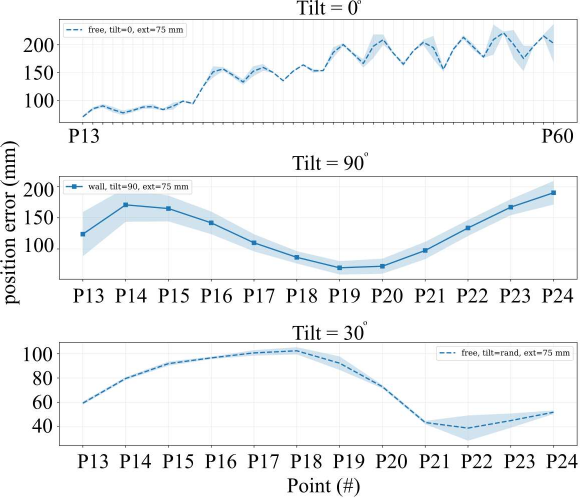}
    \caption{Variation of Euclidean error of positions of the end effector compared to the model when only the continuum section is actuated while no tilt present.}
    \label{fig:TiltConfiguration_position_error}
\end{figure}

Using these tendon lengths, we actuated the continuum section according to experiment-1 in \cref{tab:experimentDesign} and recorded the pose of each frame. At every target pose, the manipulator was held static for 10 s to collect sufficient resting data of the rigid-body poses. The procedure was repeated three times.  The end-effector trajectories from both the model and Experiment 1 were visualized in Cartesian coordinates, as illustrated in \cref{fig:continuumXYZ_positions}. We evaluated the Euclidean position error between the model and Experiment 1, together with its standard deviation, as shown in \cref{fig:TiltConfiguration_position_error}.  Throughout the experiment the Euclidean error is evaluated using the following formulas:

\begin{align}
e_{k,i}^{\text{pos}}
  &= \sqrt{(x_{k,i}-\bar{x}_{k})^{2}+(y_{k,i}-\bar{y}_{k})^{2}+(z_{k,i}-\bar{z}_{k})^{2}}, \label{eq:pos_error}\\
\mu_{k}^{\text{pos}}
  &= \tfrac{1}{N_{k}}\sum_{i=1}^{N_{k}} e_{k,i}^{\text{pos}}, \label{eq:pos_mean}\\
\sigma_{k}^{\text{pos}}
  &= \sqrt{\tfrac{1}{N_{k}}\sum_{i=1}^{N_{k}}\!\left(e_{k,i}^{\text{pos}}-\mu_{k}^{\text{pos}}\right)^{2}}. \label{eq:pos_std}
\end{align}

where $e_{k,i}^{\text{pos}}$ denotes the Euclidean position error of the $i$-th sample in slice $k$, $(x_{k,i}, y_{k,i}, z_{k,i})$ are the measured coordinates, and $(\bar{x}{k}, \bar{y}{k}, \bar{z}{k})$ are the corresponding model coordinates of slice $k$. $N{k}$ represents the total number of samples in slice $k$, $\mu_{k}^{\text{pos}}$ is the mean Euclidean position error across all $N_{k}$ samples as defined in \eqref{eq:pos_mean}, and $\sigma_{k}^{\text{pos}}$ is the standard deviation of the position error as defined in \eqref{eq:pos_std}. Here, the quaternion means are calculated using Markley's quaternion averaging \cite{markley2007quaternion}.

Informed by the results of experiment-1 where the Tilt-X is vertically down, we found that there is a huge shift in the error from below 100 mm to above 100 mm when moving from the first slice of samples to the second. Therefore, we limited the experiments 2,3 and 4 to slice-1 samples. 

Continuing with the experiments-2,3 and 4, we followed the same overall procedure of data collection under highest throttle level of the propellers, as per the combinations of conditions presented in \cref{tab:experimentDesign}.

\subsection{Flight test}
We conducted flight tests to demonstrate the feasibility of Tilt-X, as shown in \cref{fig:main_figure}. The CAAMS was flown in position-control mode, where the operator remotely initiated take-off and maintained hover without intervention. In the first test, Tilt-X executed a full cycle through the points of slice-1 in consecutive bending actuations while tilted at 90$\textdegree$ with no extension, as shown in \cref{fig:main_figure}(a),(b).

In the second test,  we emulated a real-world wall inspection scenario by mounting a plastic conduit through a hole in a flat board to create a dummy wall, as shown in \cref{fig:main_figure}(c),(d). The CAAMS was flown in position-control mode toward the conduit and hovered in close proximity based on precomputed positions. Tilt-X was then commanded to extend, to maneuver the end effector through the conduit, and retract.

\section{Results}
\label{sec:Results}

\begin{figure}
    \center
    \includegraphics[width=1\linewidth]{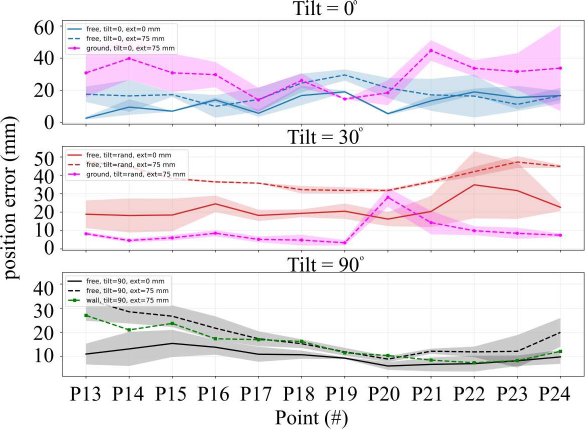}
    \caption{Position error under 100\% throttle level with reference to the mean position at no throttle. The lines represent the mean and the shading shows the standard deviation. }
    \label{fig:positionError}
\end{figure}

\begin{figure}[ht]
    \centering
    \includegraphics[width=1\linewidth]{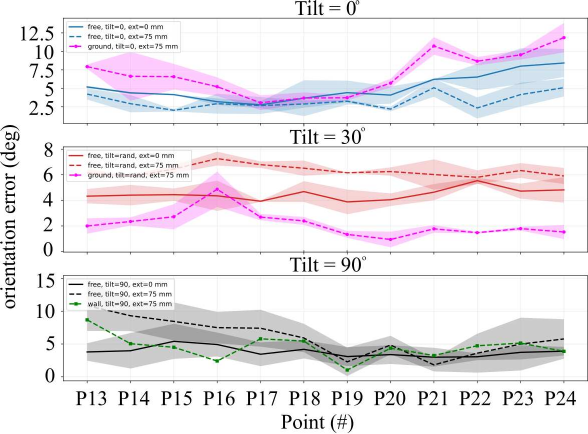}
    \caption{Orientation error under 100\% throttle level with reference to the mean orientation at no throttle. The lines represent the mean and the shading shows the standard deviation.}
    \label{fig:OrientationError}
\end{figure}

\cref{fig:continuumXYZ_positions} plots the cartesian coordinates of the end effector frame as observed in the experimental study an in comparison to the kinematic model. The corresponding errors in position are shown in \cref{fig:TiltConfiguration_position_error}.

\cref{fig:positionError} and \cref{fig:OrientationError} show the Euclidean position error and the orientation error of the bench-top experiments with sample points P13 to P24. The errors are calculated by comparing the end effector poses when the UAV propellers are at their maximum throttle level and when there is no propeller actuation present, under the different test cases.
\section{DISCUSSION}

The baseline–model discrepancy (\cref{fig:continuumXYZ_positions} and \cref{fig:positionError}) which increases with the bending is likely driven by unmodeled effects in the pure constant-curvature assumption, including backbone stiffness, tendon–routing friction, structural tolerances and the continuum section’s self-weight. Further, we evaluated downwash at maximum throttle as a worst-case condition; in practical deployments, a CAAMS operating with available payload margin would typically not require full thrust and therefore would experience lower downwash levels. 

When tilt is 0$^\circ$, and the system is away from the ground, position error is least when the telescoping structure is retracted (\cref{fig:positionError}(a)). The error increases with extension in the telescoping structure. When the drone approaches the ground in the extended case, the error increases further. In both the extended cases the standard deviation is higher than the retracted case. This observation is likely due to increased turbulence near the ground and air bouncing off the surface causing oscillations in the manipulator position \cite{david2024ground, bartholomew2014learning}.

When tilt is 30$^\circ$, the standard deviation of error reduces in the extended position (\cref{fig:positionError}(b)), suggesting that the manipulator has escaped the influence of downwash. The error is least when ground effect present, and shows less fluctuation than the free condition. A thorough aerodynamic study in the future would explain the reason for this. 

The behaviour when tilt is 90$^\circ$ (\cref{fig:positionError}(c)) is similar to the above case. Once again, the lower deviation in error in both the extended cases is a sign of the manipulator leaving the influence zone of the propeller induced disturbances.

Observing \cref{fig:OrientationError}, orientation errors show a trend similar to that of position error. In both position and orientation, the error increases around points P22 to P24. This is due to asymmetry in the construction of the UAV, where the onboard computer is in the path of the air between the propellers and the manipulator, likely leading to turbulence. 

It should be noted that the analysis considers the Euclidean error in position and the combined error in orientation. Further study of the individual components would shed light on the direction of motion or bending that is most causing the error, and assist in improvements to design and control.

In flight tests (\cref{fig:main_figure}), the system successfully demonstrated the deployment of the CM through a conduit placed horizontally. Manipulation in this region in front of the UAV is made possible because of the Tilt-X mechanism, and showcases the benefit of its design.  
\section{conclusions}
In this paper, we present a novel design for a continuum arm aerial manipulation system. We showcase a design and its kinematics, which uniquely positions all actuators at fixed locations on the UAV while being capable of tilting, extending and bending the continuum manipulator. Comparisons of experimental results with an ideal model showcase similar trends with increasing deviations for more pronounced bends of the continuum portion. Our study also characterized the effect of downwash and wind effects when in proximity to the ground and a wall. Results show that the manipulator is stabilized when it extends out of the zone of influence of downwash. 

The present study is empirical and an exact representation of the manipulator in reality. An aerodynamic simulation or visualisation of fluid flow around the manipulator would complement the understanding of downwash and is being considered in the future. Our method presents a systematic way to characterise the behaviour of a continuum manipulator, the data from which can be used to build a model of the system that more closely represents actual behaviour than a simulation. Currently, our kinematics is open-loop and such a model is expected to be useful when designing a feedback controller. Future work will focus on strengthening the continuum section to increase payload capability and developing feedback control for precision manipulation.
\bibliographystyle{IEEEtran} 

% --- tighten bibliography spacing (IEEEconf) ---
\makeatletter
\let\oldthebibliography\thebibliography
\let\endoldthebibliography\endthebibliography
\renewenvironment{thebibliography}[1]{%
  \begin{oldthebibliography}{#1}%
  \setlength{\itemsep}{0pt}%
  \setlength{\parskip}{0pt}%
  \setlength{\parsep}{0pt}%
}{%
  \end{oldthebibliography}%
}
\makeatother

\bibliography{references}

\end{document}